\documentclass[letterpaper, 10pt, conference]{ieeeconf}      % Use this line for a4 paper

\IEEEoverridecommandlockouts                              % This command is only needed if 
\usepackage{amsmath} % assumes amsmath package installed
\usepackage{graphicx}
\usepackage{placeins}
\usepackage{amsfonts}
\usepackage{dblfloatfix} % Enables double-column floats at the bottom of the page
\usepackage{pifont}% http://ctan.org/pkg/pifont
\newcommand{\cmark}{\ding{51}}%
\newcommand{\xmark}{\ding{55}}%

\DeclareMathOperator{\softmax}{softmax}

\title{\LARGE \bf
ZS6D: Zero-shot 6D Object Pose Estimation using Vision Transformers
}

\author{Philipp Ausserlechner$^{1}$, David Haberger$^{1}$, Stefan Thalhammer$^{1}$, Jean-Baptiste Weibel$^{1}$ and Markus Vincze$^{1}$% <-this % stops a space
\thanks{*This work was supported by the EU-program EC Horizon 2020 for Research and Innovation under grant agreement No. 101017089, project TraceBot.}% <-this % stops a space
\thanks{$^{1}$All authors are with Vision for Robotics Laboratory, 
Automation and Control Institute, 
TU Wien, Austria {\tt\small \{ausserlechner, haberger, thalhammer, weibel, vincze\}@acin.tuwien.ac.at}}%
}

\begin{document}

\maketitle
\thispagestyle{empty}
\pagestyle{empty}

%%%%%%%%%%%%%%%%%%%%%%%%%%%%%%%%%%%%%%%%%%%%%%%%%%%%%%%%%%%%%%%%%%%%%%%%%%%%%%%%
\begin{abstract}
As robotic systems increasingly encounter complex and unconstrained real-world scenarios, there is a demand to recognize diverse objects.
The state-of-the-art 6D object pose estimation methods rely on object-specific training and therefore do not generalize to unseen objects.
Recent novel object pose estimation methods are solving this issue using task-specific fine-tuned CNNs for deep template matching.
This adaptation for pose estimation still requires expensive data rendering and training procedures. 
MegaPose for example is trained on a dataset consisting of two million images showing 20,000 different objects to reach such generalization capabilities.
To overcome this shortcoming we introduce ZS6D, for zero-shot novel object 6D pose estimation.
Visual descriptors, extracted using pre-trained Vision Transformers (ViT), are used for matching rendered templates against query images of objects and for establishing local correspondences.
These local correspondences enable deriving geometric correspondences and are used for estimating the object's 6D pose with RANSAC-based P\textit{n}P.
This approach showcases that the image descriptors extracted by pre-trained ViTs are well-suited to achieve a notable improvement over two state-of-the-art novel object 6D pose estimation methods, without the need for task-specific fine-tuning.
Experiments are performed on LMO, YCBV, and TLESS.
In comparison to one of the two methods we improve the Average Recall on all three datasets and compared to the second method we improve on two datasets.
\end{abstract}

%%%%%%%%%%%%%%%%%%%%%%%%%%%%%%%%%%%%%%%%%%%%%%%%%%%%%%%%%%%%%%%%%%%%%%%%%%%%%%%%
\section{INTRODUCTION}

Robotics, and service robotics, in particular, have the potential to profoundly transform our society.
However, enabling semantic manipulation requires estimating the poses of objects, which presents substantial challenges for a constantly increasing set of objects.
Contemporary pose estimation methods~\cite{labbe2020cosypose,park2019pix2pose,su2022zebrapose,wang2021gdr,thalhammer2023cope} are trained for specific objects and do not generalize to unseen ones.
Their little flexibility and adaptability require re-training every time the set of objects that need to be handled by the robot changes.

\begin{figure}[t!]
  \centering
  \includegraphics[width=0.48\textwidth]{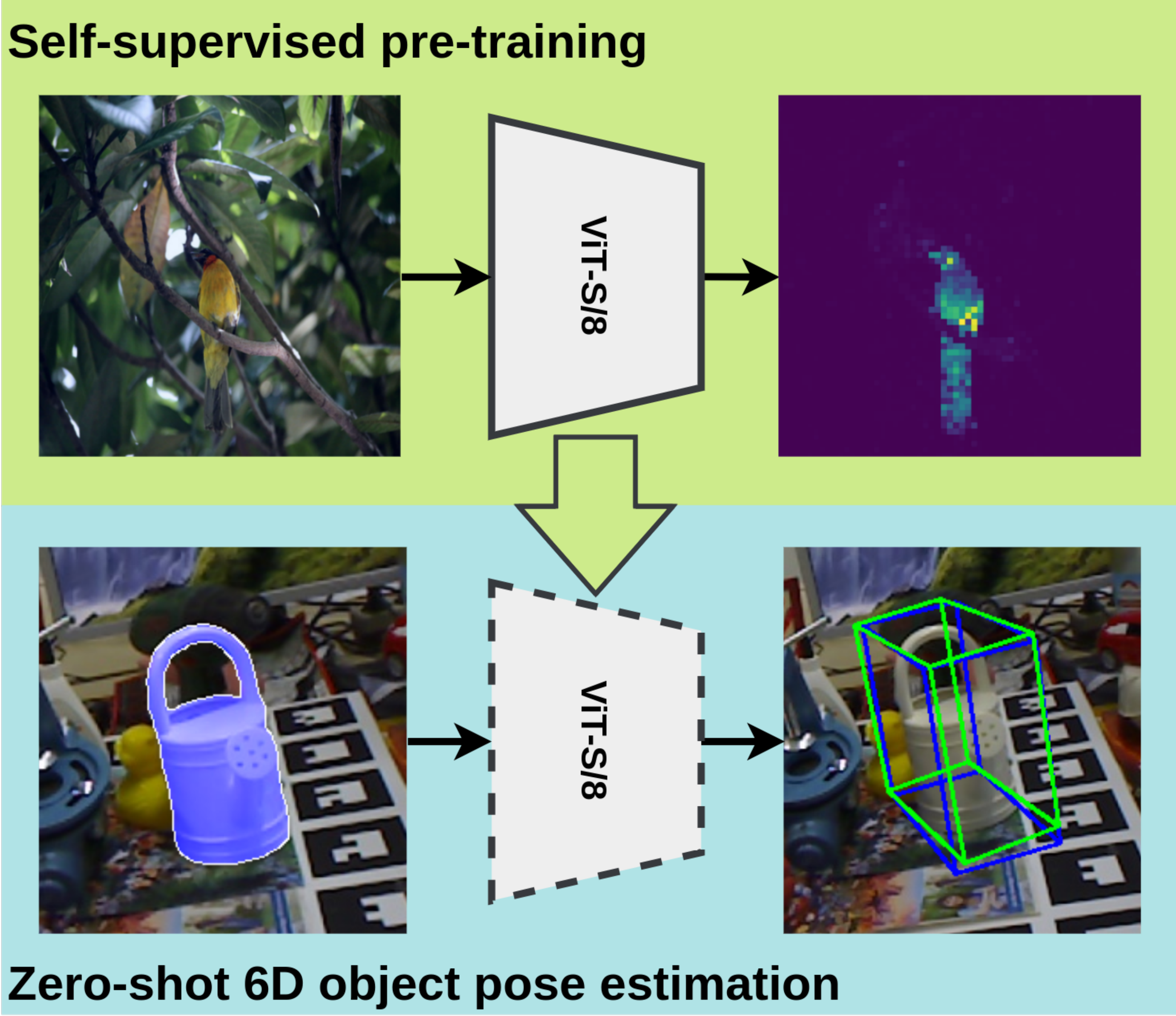}
  \caption{\textbf{Zero-shot 6D pose estimation} Descriptors produced by a self-supervised ViT~\cite{caron2021emerging} are descriminative enough for novel object \(6D\) pose estimation, without any task-specific fine-tuning in a zero-shot manner.}
  \label{fig:teaser}
\end{figure}

Recent novel object pose estimation approaches provide a feasible solution to this problem by matching query images against rendered templates of the object models~\cite{shugurov2022osop,labbe2022megapose,nguyen2022templates,thalhammer2023self}.
Such deep template matching requires task-specific fine-tuning.
Diverse object models are used for rendering training data, with e.g. BlenderProc~\cite{denninger2020blenderproc}, which is used for training multi-staged CNN pipelines.
In the case of MegaPose~\cite{labbe2020cosypose}, two million scene-level training images featuring $20,000$ different object instances are rendered. 
These images are used for training their deep template matcher.
Using such strategies partially alleviates the need for object-specific training, yet assumes that the set of training objects is enough to generalize to arbitrary real-world objects. 
This strategy becomes untractable to obtain a system that handles all objects. 

In this work we hypothesize that self-supervised pre-trained Vision Transformers~\cite{dosovitskiy2020image,caron2021emerging} (ViT) are bound to overcome the requirement of task-specific fine-tuning, since recent works indicate the generality of their extracted descriptors~\cite{amir2021deep,caron2021emerging,oquab2023dinov2,thalhammer2023self}.
In order to verify our hypothesis we present ZS6D, Figure~\ref{fig:teaser}, a method for zero-shot $6D$ object pose estimation.
Our method extracts image descriptors to match a query image against rendered object templates.
Subsequently, local correspondences between the query and the matched template are computed to derive geometric correspondences and estimate the pose using RANSAC-based~\cite{fischler1981random} P\textit{n}P~\cite{moreno2007accurate}. 
In practice, we use colored object coordinates, i.e. object vertex locations mapped to RGB values~\cite{park2019pix2pose}.
%, which are used together with RANSAC-based P\textit{n}P to compute the object's pose.
As these are defined in object space and derived using the matched local correspondences, the retrieved pose grants a higher accuracy than the available templates provide.
This allows our approach to overcome noise in the template matching and achieve accurate object poses with as few as $200$ object templates.
We provide experiments showing that using ViTs for zero-shot $6D$ object pose estimation alleviates the requirement for both training data and model fine-tuning. 
Besides, ZS6D also improves the Average Recall (AR)~\cite{hodan2018bop} on two of three tested standard datasets, in comparison to the state of the art.
%We generate templates from uniformly distributed views and compare them with segmented objects.
%We then identify the best matching template and establish local correspondences between the query image and the template. 
%These point pairs are subsequently lifted to 2D-3D correspondences using the template's uv map. 
%Finally, the \(6D\) pose of the objects is obtained through a Perspective-n-Point algorithm (P\textit{n}P)~\cite{moreno2007accurate} with RANSAC~\cite{fischler1981random} iterations.
Our contributions to the field of \(6D\) object pose estimation are the following:
\begin{itemize}
    \item We present a pose estimation method that estimates $6D$ object poses in a zero-shot fashion. The presented methods improve over the state of the art for novel object pose estimation on two standard datasets using the Average Recall~\cite{hodan2018bop}.
    \item We demonstrate that pre-trained Vision Transformers (ViT) improve over task-specific fine-tuned CNNs for novel object \(6D\) pose estimation. %The presented results  that ViTs improve over two fine-tuned CNNs on two benchmark datasets
    %\item We demonstrate that non-fine-tuned self-supervised ViTs (foundational models) are suitable for \(6D\) object pose estimation and outperform fine-tuned CNNs on two benchmark datasets
\end{itemize}

%In contrast, we solely rely on the descriptors of ImageNet1k-pretrained~\cite{russakovsky2015imagenet} Vision Transformers~\cite{dosovitskiy2020image} (ViT) without fine-tuning them, i.e. using them in a zero-shot fashion.
%The ViT is trained in a self-supervised manner, which enables its application in a variety of computer vision tasks and therefore can be described as a foundational Computer Vision model~\cite{caron2021emerging}.
%Such models also exhibit higher robustness against dataset biases.
%Their broad learning and application potential have emerged as a key driver of progress in fields such as Natural Language Processing. 
%In parallel, the rise of self-supervised Vision Transformers~\cite{caron2021emerging} in Computer Vision has shown promising results in various tasks, including object classification, segmentation, and image retrieval. 
%Recent publications suggest that ViT-based foundational models are more general than CNNs for template matching \cite{thalhammer2023self}.
%In the domain of single reference pose estimation, where one annotated real-world sample is used for correspondence matching foundational models are already successfully applied for 3D pose estimation~\cite{goodwin2022zero,fan2023pope}.
%In terms of practical application, a 3D pose is not sufficient for enabling robotic interaction.
%Our method obtains the \(6D\) pose of unseen objects without fine-tuning or annotating any real-world samples, by utilizing object templates in combination with the corresponding image descriptors generated by a foundational model~\cite{caron2021emerging}.

The paper proceeds as follows: we present relevant state-of-the-art methods in Section~\ref{sec:rel_works}, our proposed evaluation scheme in Section~\ref{sec:method}, and our experimental results in Section~\ref{sec:results} before presenting our conclusions in Section~\ref{sec:conclusion}.

%%%%%%%%%%%%%%%%%%%%%%%%%%%%%%%%%%%%%%%%%%%%%%%%%%%%%%%%%%%%%%%%%%%%%%%%%%%%%%%%
\section{Related Work} \label{sec:rel_works}
This section presents the state of the art for \(6D\) object pose estimation with the main focus on novel object pose estimation. 
Following that self-supervised Vision Transformers as discussed.

Contemporary methods for pose estimation~\cite{wang2021gdr,su2022zebrapose,labbe2020cosypose,park2019pix2pose,liu2021kdfnet, thalhammer2021pyrapose, lin2022single, richter2021handling, jawaid2023towards} rely on object-specific training and a preceding object detection stage which also has to be trained separately.
These approaches do not scale well, since they have to be trained for every new object.
In contrast, \cite{zakharov2019dpod, thalhammer2023cope, hodan2020epos}~scale better because they are trained for an entire set of objects simultaneously, integrating object detection and pose estimation in a single stage.
Nevertheless, all of these methods lack practicality in many real-world scenarios, since it is not feasible to re-train for every new set of objects encountered.
Recent single reference image pose estimation methods like Pope~\cite{fan2023pope} and Goodwin et al.~\cite{goodwin2022zero} leverage the descriptors produced by self-supervised ViTs to estimate the relative rotation between the reference image and the detected object.
However, these approaches are insufficient for robotic applications, since a \(6D\) pose is required for object manipulation. 

\textbf{Novel object pose estimation:}
We refer to the problem of estimating the pose of unseen objects during training as novel object pose estimation.
A classical approach to this problem is the Point Pair Feature (PPF) method~\cite{drost2010ppf}. 
It leverages depth information by approximating local geometries of the query image and uses it as a hash to match the object model.
DeepIM~\cite{li2018deepim} is one of the first approaches that leveraged CNN-based features to iteratively refine the pose of a template compared to the query image.
Another noteworthy step towards novel object pose estimation comes from Sundermeyer et al.~\cite{sundermeyer2020multi}, which uses a common encoder that generalizes to unseen objects and extracts descriptive image features.
Ngyuen et al.~\cite{nguyen2022templates} revisits the idea of template matching by applying CNN-based features to estimate the rotation of unseen objects from query images.
Thalhammer et al.~\cite{thalhammer2023self} extends this scheme and demonstrates that ViTs outperform CNNs for template matching.
With the exception of DeepIM, these approaches only estimate a rotation, which is not sufficient for robotic interaction.
More recent methods like OSOP~\cite{shugurov2022osop} deploy a task-specific fine-tuned CNN to derive dense correspondences between the query image and a large set of templates, \(5K\) in the case of LMO.
Another noteworthy approach is MegaPose~\cite{labbe2022megapose} which relies on an initial template-matching followed by an iterative refinement.
They use a CNN which is trained on a large-scale dataset with more than \(20,000\) objects and two million images, which allows them to effectively generalize deep template matching to unseen objects.
Our method differs from these approaches by relying solely on a self-supervised pre-trained ViT, with no requirement for pose estimation-specific fine-tuning and a comparably small set of templates (up to $300$).

\begin{figure*}[t]
  \centering
  \includegraphics[width=\textwidth]{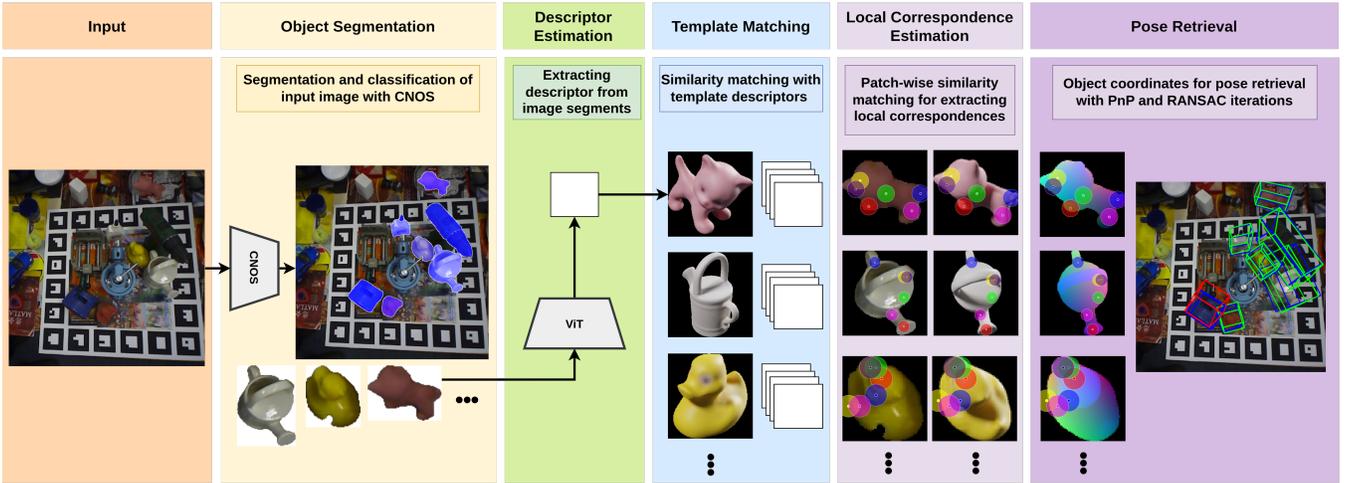}
  \caption{\textbf{Overview of ZS6D} The diagram depicts the stages of the ZS6D pose estimation pipeline. Initially, Segment Anything~\cite{kirillov2023segment} segmentations~\cite{nguyen2023cnos} are used to isolate the object of interest. Then, dense visual descriptors are extracted~\cite{amir2021deep} from the segmented object, followed by a comparison against pre-rendered template descriptors using cosine similarity. The image further illustrates the process of matching local correspondences between the selected template and the segmented region, which enables the derivation of 2D-3D correspondences from the template's colored object coordinates. The final step is the application of a P\textit{n}P~\cite{moreno2007accurate} algorithm with RANSAC~\cite{fischler1981random} iterations to obtain the \(6D\) object pose.}
  \label{fig: pipeline}
\end{figure*}

\textbf{Vision Transformers:}
In natural language processing, transformers~\cite{vaswani2017attention} are the dominant architecture due to their capability to be trained on large-scale datasets in a self-supervised manner. 
Many efforts were made to transfer this architecture to the Vision domain~\cite{ramachandran2019stand}, resulting in the Vision Transformer (ViT)~\cite{dosovitskiy2020image}. 
These models perform comparably better than CNNs, but their advantages really materialize with self-supervised training.
This procedure allows them to generalize well to novel tasks and makes them robust against dataset biases~\cite{caron2021emerging}. 
Such foundational Computer Vision models show comparable results to the state of the art for supervised models in tasks like object classification, segmentation~\cite{kirillov2023segment}, and image retrieval. 
Latest publications~\cite{goodwin2022zero,fan2023pope,thalhammer2023self} show that ViTs can be applied without fine-tuning for object pose estimation, with respect to a reference image to estimate a \(3D\) pose.
We show in our experiments that those foundational Computer Vision models can be applied to obtain the full \(6D\) pose of unseen objects without any fine-tuning.

\section{ZS6D} \label{sec:method}
In this section, we propose our method for zero-shot \(6D\) object pose estimation, named ZS6D, which solely relies on an object model and the descriptors generated by a self-supervised ViT.
Figure~\ref{fig: pipeline} provides an abstract visualization of the pose inference.
In the following subsections, we describe how the objects are segmented in the query image, how the best matching template is selected, and how the local correspondences are obtained.

\subsection{Object Detection and Segmentation}
ZS6D assumes the availability of object instance segmentation masks and a query image \(I_s\).
The segmentations are generated with the zero-shot \(2D\) approach of CNOS~\cite{nguyen2023cnos}.
When looking at a new scene, the Segment Anything Model (SAM) of~\cite{kirillov2023segment} is employed to generate object segmentation proposals denoted as \(\{I_{p} \mid  I_p \subseteq I_S\}\).
The descriptors \(D_p^{seg}\) of each object proposal as well as the template descriptors \(D_t^{seg}\) are generated by a single forward pass through a ViT~\cite{oquab2023dinov2}.
The cosine similarities between all the template descriptors \(D_t^{seg}\) and \(D_p^{seg}\) are calculated to recognize the object within the proposal, after aggregating the similarity scores by object class. 
The class of the object proposal is determined by the highest aggregated score.

\subsection{Global Descriptor Estimation} \label{sec:desc}
To estimate the image descriptors we apply a self-supervised ViT~\cite{caron2021emerging}. 
The core operation of a ViT~\cite{dosovitskiy2020image} is the attention mechanism~\cite{ramachandran2019stand}.
We define the input image \( X \in \mathbb{R}^{n \times d} \) as a sequence of patches \( (x_{1}, x_{2}, \ldots, x_{n}) \). 
The aim of self-attention is to estimate the interaction between all \( n \) patches. 
Therefore, we define three learnable weight matrices: \(\mathbf{W}^Q \in \mathbb{R}^{d \times d_q}\), \(\mathbf{W}^K \in \mathbb{R}^{d \times d_k}\), and \(\mathbf{W}^V \in \mathbb{R}^{d \times d_v}\). These matrices allow us to transform the input sequence \( \mathbf{X} \) into Queries \( \mathbf{Q} = \mathbf{X}\mathbf{W}^{\mathbf{Q}}\), Keys \( \mathbf{K} = \mathbf{X}\mathbf{W}^{\mathbf{K}}\), and Values \( \mathbf{V} = \mathbf{X}\mathbf{W}^{\mathbf{V}}\) respectively. The self-attention is then computed as:

\begin{align}
Z = \softmax(\frac{\mathbf{Q}\mathbf{K}^\top}{\sqrt{d_q}})\mathbf{V}
\end{align}

The ViT itself consists of multiple self-attention layers, therefore creating multiple options for choosing a viable image descriptor.
For example, CNOS~\cite{nguyen2023cnos} uses the class token which is a vector that gets passed together with the image patches through the network and serves as a global image embedding. 
In~\cite{thalhammer2023self}, the authors show that patch-wise token embeddings are more suitable for 3D pose estimation than the class token. 
Furthermore, the authors of~\cite{amir2021deep} empirically show that the key token \(K\) embedding from layer 9 of the ViT is the most suitable for global image description. 
The authors argue that the shallow layers are the best to represent global geometric information. We follow their argumentation and use these as image descriptors.  

\subsection{Template Matching}
The descriptor of the query image \(D_p\) is compared against a set of template descriptors \(\{D_t, \, \forall t \in T\}\), both created by a single forward pass through the ViT.
Similar to classical approaches we assume uniform coverage of the viewing space. 
Thus we rely on rendered object views~\cite{aldoma2011cad,hodavn2015detection}.
In Section~\ref{sec:results}, we present detailed ablations justifying the $300$ uniformly distributed views we chose to ensure comprehensive coverage of the object model.
To estimate the closest template we compute the cosine similarity between the descriptor of the object proposal and each descriptor from the set of templates, according to:

\begin{align}
\max{\langle D_{t}, D_{p}\rangle}, \forall t \in T 
\end{align}

The template with the highest similarity score is used for estimating local correspondences.

\subsection{Local Correspondence Estimation} \label{sec:geo_corrs}
The matched template provides a coarse pose estimate with the maximum accuracy limited by the view coverage of the object. 
As an example, directly retrieving the rotation of the input object requires a large number of templates, e.g. \(21,672\) templates for TLESS~\cite{nguyen2022templates}. 
Furthermore, retrieving the object's translation using the ratio of the estimated bounding box to the rendered template depends on the rotational error between the query image and template, thus translation error increases with rotation error.
In order to circumvent these issues, we estimate and match local correspondences between query and template images.
ViTs~\cite{dosovitskiy2020image} treat images as local patches and estimate relations between these, to obtain local descriptors. 
We aim to match corresponding patches between query \(I_{p}\) and the template images \(\{I_{t}\}\).
For this purpose we adopt the key \(k\) token from layer 11 as our patch descriptor, a deeper layer found to yield superior performance when estimating local correspondences~\cite{amir2021deep}. 
Here, \(Q = \{q_i\}\) represents descriptors of the template, while \(P = \{p_i\}\) denotes descriptors from the segmented region of the query image. 
To identify the optimal correspondences, we compute the nearest neighbor \(NN\) according to the cosine similarity for each descriptor in \(Q\) and \(P\), retaining only those correspondences, termed \(LC\), where the patch demonstrates the highest congruence in both directions:

\begin{align}
\begin{split}
LC(Q,P) = \{&(q,p) \, | \, q \in Q, \, p \in P, \\
            &NN(p,Q) = q \, \land \, NN(q, P) = p \}
\end{split}
\label{eq:local_corr_calc}
\end{align}

\subsection{Pose Retrieval} 
From the correspondence estimation stage, only robust patch pairs are left. 
Given that we know the pose of the object of interest in the rendered templates, we can obtain the corresponding 3D coordinates in object space by looking up the values of the templates' colored object coordinates. 
Each local correspondence, therefore, yields colored object coordinates, that we use in the P\textit{n}P~\cite{moreno2007accurate} algorithm with RANSAC iterations~\cite{fischler1981random} to recover the final \(6D\) pose of the segmented target object from the query image \(I_{p}\).

\section{Experiments} \label{sec:results}
In this Section, we discuss the experimental setup. 
We compare our method to the state of the art for novel object \(6D\) pose estimation on three of the core datasets of the Benchmark for \(6D\) Object Pose Estimation challenge~\cite{hodan2018bop} (BOP). Additionally, we provide ablation studies evaluating the impact of the segmentation quality, as well as
selecting the optimal number of views for template generation, and the optimal number of local correspondences for object coordinate estimation.

\begin{figure}[t]
  \centering
  \includegraphics[width=0.45\textwidth]{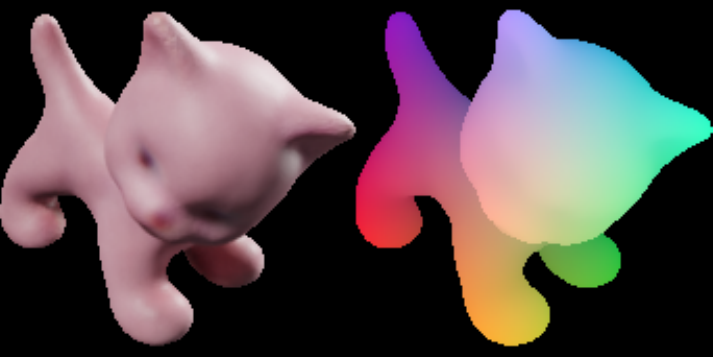}
  \caption{\textbf{Template and corresponding object coordinates} Visualization of a template of LMO's cat and the corresponding colored object coordinates.}
  \label{fig:templ_uv}
\end{figure}

\subsection{Datasets}
We evaluate our ZS6D on three of the core BOP datasets \cite{hodan2018bop}, LMO \cite{brachmann2014learning}, YCBV \cite{xiang2017posecnn}, and TLESS \cite{hodan2017t}. 
These three datasets reflect standard challenges for object pose estimation, occlusion in the case of LMO, strong illumination changes for YCBV, and texture-less objects for TLESS. 
Since our method infers poses in a zero-shot fashion, we do not require the training sets and for testing the respective test sets as they are used in BOP.

\subsection{Implementation Details}
BlenderProc~\cite{denninger2020blenderproc} is used for rendering templates since it is considered the standard tool for that purpose~\cite{labbe2022megapose,nguyen2022templates,thalhammer2023self}.
For each object, we uniformly sample views on a regular icosahedron. 
We use $300$ templates per object unless stated otherwise.
Colored object coordinates are used as geometric correspondences, Figure~\ref{fig:templ_uv}, for pose retrieval with P\textit{n}P~\cite{park2019pix2pose}.

For global descriptor estimation (Section~\ref{sec:desc}) and local correspondence estimation (Section~\ref{sec:geo_corrs}), \(ViT-S/8\)~\cite{caron2021emerging} is used, where \(8\) refers to number of pixels for each side of the patches. 
We use the weights pre-trained on ImageNet1k~\cite{russakovsky2015imagenet}.
The input image resolution to both stages is \(224 \times 224\).
A descriptor size of $384$ and $6528$ is used for global descriptor and local correspondence estimation, respectively.
Small patch sizes are crucial for estimating meaningful local correspondences in order to robustly match corresponding patches between query and template images using nearest neighbors.
All experiments report the Average Recall \(AR\) metric of the BOP~\cite{hodan2018bop}, which is the standard for \(6D\) object pose estimation.

\subsection{Object Segmentation}
For all presented experiments we use the segmentation masks provided by CNOS~\cite{nguyen2023cnos} unless stated otherwise.
The templates for classifying the SAM masks~\cite{kirillov2023segment} are rendered from the provided object models. 
As proposed by the authors of CNOS, \(V=42\) viewpoints on a regular icosahedron are used for generating templates to ensure a uniformly distributed view coverage of the object. 
Subsequently, template descriptors \(D_t\) are computed using DINOv2~\cite{oquab2023dinov2}. 
The class token is used as descriptor \(D_t\) and its dimension is \(N_O \times V \times C\). 
We follow their hyperparameter configuration of \(C=1024\). 

\begin{table}[t!]
\begin{center}
\begin{tabular}{|l|c|c|c|c|}
\hline
Method & Zero-shot &  LMO & YCBV & TLESS \\
\hline\hline
MegaPose~\cite{labbe2022megapose} & \xmark & 0.187 & 0.139 & 0.197 \\ 
OSOP~\cite{shugurov2022osop} & \xmark & 0.274 & 0.296 & \textbf{0.403} \\ \hline
%OSOP~\cite{shugurov2022osop} & \xmark & $n$ & 0.312 & 0.332 & - \\ \hline
ZS6D (\textbf{Ours}) & \cmark & \textbf{0.298} & \textbf{0.324} & 0.210 \\
%ZS6D (\textbf{Ours}) & \cmark & $n$ & \textbf{0.369} & \textbf{0.378} & 0.225 \\
\hline
\end{tabular}
\end{center}
\label{tab:main_results}
\caption{\textbf{Evaluation for unseen objects} Results are provided using the Average Recall ($AR$) of BOP. We compare against the initial pose estimates from MegaPose~\cite{labbe2022megapose} and OSOP~\cite{shugurov2022osop} since we only use RGB input and do not use a refinement stage.}
\end{table}

\begin{table}[t!]
\begin{center}
\begin{tabular}{|l|c|c|c|c|}
\hline
Method & Mask Origin & LMO & YCBV & TLESS \\
\hline\hline
ZS6D (\textbf{Ours}) & CNOS~\cite{nguyen2023cnos} & 0.298 & 0.324 & 0.210 \\
ZS6D (\textbf{Ours}) & ground truth & 0.527 & 0.499 & 0.460 \\
\hline
\end{tabular}
\end{center}
\label{tab:results_gt}
\caption{\textbf{Influence of segmentation masks} Pose estimation results with CNOS~\cite{nguyen2023cnos} and ground truth masks. Using a single pose hypothesis for evaluation.}
\end{table}

\begin{figure*}[t]
  \centering
  \includegraphics[width=\textwidth]{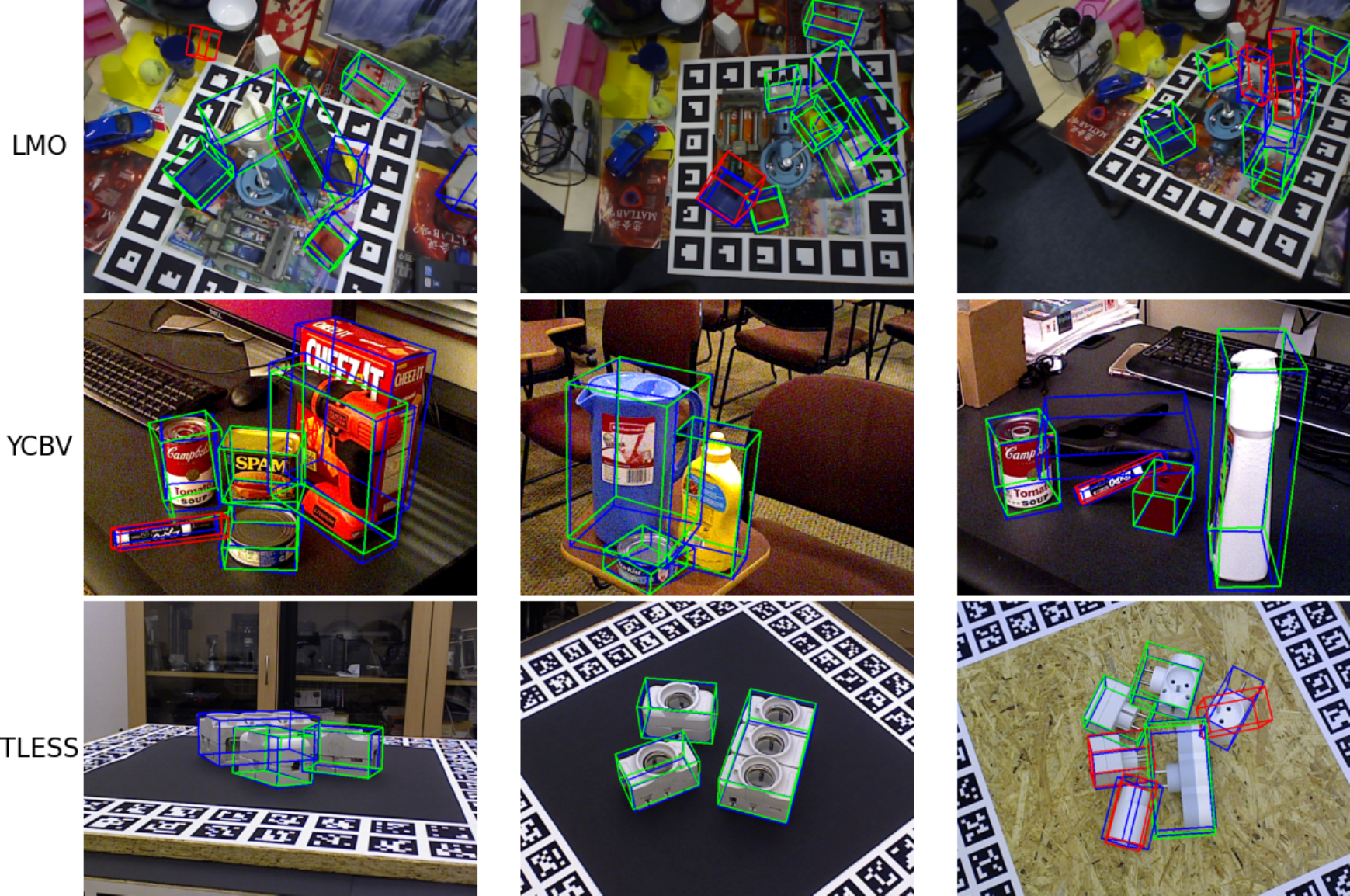}
  \caption{\textbf{Qualitative results} Visualized are object poses as $3D$ bounding boxes, blue indicates ground truth, green true positives, and red false positives on LMO, YCBV, and TLESS.}
  \label{fig:qualitative}
\end{figure*}

\subsection{Main Results}
This section presents our main results for zero-shot novel object \(6D\) pose estimation. 
A comparison is provided against MegaPose~\cite{labbe2022megapose} and OSOP~\cite{shugurov2022osop}.
Table~\ref{tab:main_results} reports results for pose initialization without a refinement stage using RGB as input are reported since this is the case for our method. 
MegaPose and OSOP apply task-specific fine-tuned CNNs for object pose estimation, while our results are obtained without any fine-tuning, using a pretrained \(ViT-S/8\) in a zero-shot manner.

We improve the \(AR\) on all three datasets compared to Megapose, despite it using a larger number of templates (\(520\) compared to \(300\) for our method) and relying on detections of Mask R-CNN~\cite{he2017mask}, trained on the synthetic physically-based rendered (PBR) data of the target objects.
The relative improvement is \(59\%\) on LMO, \(133\%\) on YCBV, and \(7\%\) on T-LESS.

% Evaluating against OSOP~\cite{shugurov2022osop}, we improve the \(AR\) on LMO by \(9\%\) and on YCBV by \(9\%\), but the relative performance decrease by \(48\%\) on TLESS.
% Table~\ref{tab:results_gt} shows significantly improved AR of our methods on TLESS (\(120\%\)) when using the ground truth masks, which suggests that the segmentation masks generated by CNOS~\cite{nguyen2023cnos} are less accurate for TLESS than for LMO and YCBV.
% Additionally estimating the patch-wise local correspondences on texture-less objects exhibiting symmetries leads to ambiguities.
% OSOP in contrast proposes a custom segmentation stage that matches the observations against object templates.
% We show qualitative~\ref{fig:qualitative} results to give an impression of our pose estimation pipeline on CNOS segmentation masks.

Evaluating against OSOP~\cite{shugurov2022osop}, we improve the \(AR\) on LMO and YCBV for a single hypothesis and multiple hypotheses.
On TLESS, OSOP reports a higher \(AR\) score.
Table~\ref{tab:results_gt} shows significantly improved AR of our methods on TLESS when using the ground truth masks, which suggests that the segmentation masks generated by CNOS~\cite{nguyen2023cnos} are less accurate for TLESS than for LMO and YCBV.
Additionally estimating the patch-wise local correspondences on texture-less objects exhibiting symmetries leads to ambiguities.
OSOP in contrast proposes a custom segmentation stage that matches the observations against object templates.
We show qualitative results to give an impression of our pose estimation pipeline on CNOS segmentation masks, see Figure~\ref{fig:qualitative}. 
In most of the cases, poses are correctly (green) estimated. 
In some cases, the segmentation mask is missing (no green or red box), especially for YCBV and TLESS.
The influence of the segmentation masks on the pose estimates is discussed in the next section.

\subsection{Ablations}
In this section, we present ablations to further investigate the contributing factors to the methods' performance. 
We conduct three central ablation studies to determine the impact of the segmentation masks, the number of views for template generation, and the number of local correspondences to extract for sub-sequential pose retrieval.

\subsubsection{Mask Quality}
We evaluate our ZS6D with the CNOS and compare it to the ground truth masks in order to disentangle the pose estimation accuracy of ZS6D from the detection stage. 
Results for LMO, YCBV, and TLESS are provided in Table~\ref{tab:results_gt}. 
The respective improvements using the ground truth masks are $77\%$, $54\%$, and $119\%$. 
These results indicate that large improvements are to be expected when obtaining more accurate segmentation masks as input to the presented method.
Especially for TLESS, the $AR$ score when using CNOS masks is far off the theoretically obtainable upper bound.
%This serves as an upper bound for the performance and shows that all 3 datasets have a very similar \(AR\) score. 
%This supports our claim that the performance decrease on TLESS depends on the segmentation masks of CNOS~\cite{nguyen2023cnos}. 
%Additionally, it also suggests that our method is general as it can be applied to a wide range of different scenarios with similar expected performance. 

\begin{figure}[t!]
  \centering
  \includegraphics[width=0.5\textwidth]{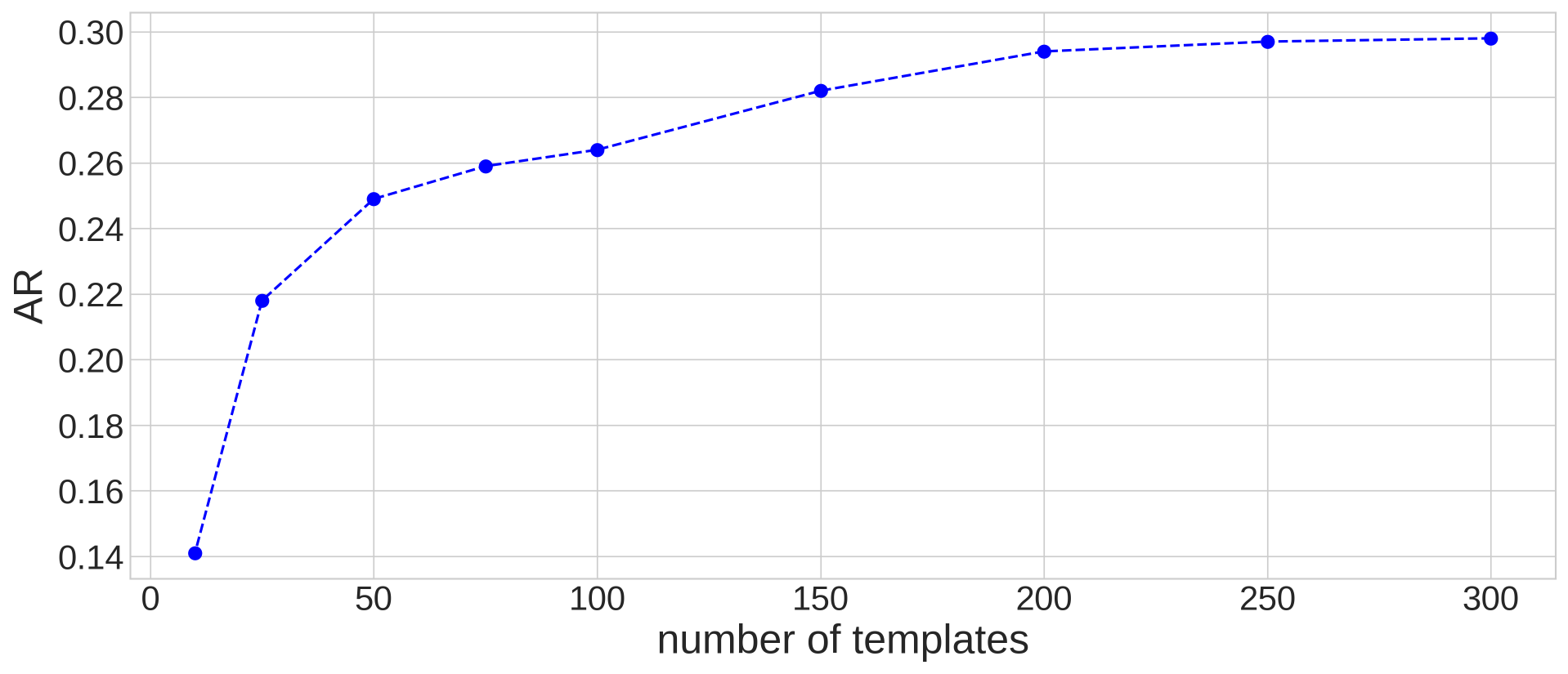}
  \caption{\textbf{Number of templates} Impact of the number of templates per object. Reported is the $AR$ score on LMO.}
  \label{fig:number_templates}
\end{figure}

\subsubsection{Number of Templates}
\begin{figure}[t]
  \centering
  \includegraphics[width=0.5\textwidth]{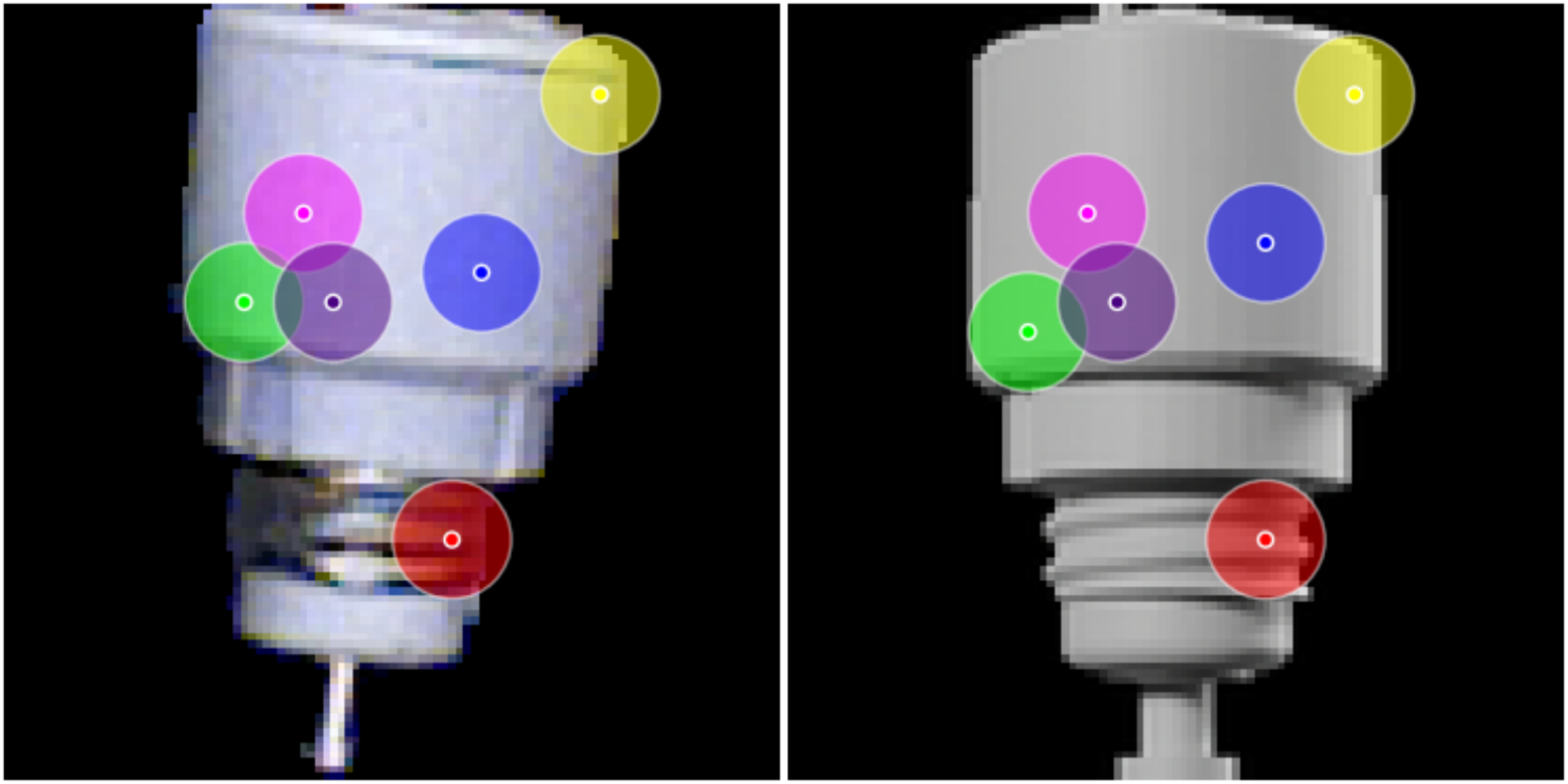}
  \caption{\textbf{Local correspondences} Visualization of matched correspondences of the query image (left) and the matched template (right). Matching colors corresponding to the same local correspondence.}
  \label{fig:correspondences_tless}
\end{figure}

%In the case of the number of templates, we observe that the AR score increases the more views we generate \ref{fig:number_templates}. 
Figure~\ref{fig:number_templates} ablates the influence of the number of templates reporting the $AR$ score on LMO.
The instance segmentation masks generated using CNOS are used as location priors.
A significant increase in accuracy is observable up to $200$ templates. 
Since ZS6D derives colored object coordinates based on local correspondence matching the retrieved pose grants a higher accuracy than the available templates provide, as indicated in Figure~\ref{fig:correspondences_tless}.
The accuracy is asymptotically approaching a maximum at $300$ views.
%, as our procedure only needs a reasonably and not a perfectly matching template, due to the subsequent local correspondence matching \ref{fig:correspondences_tless}.

% \FloatBarrier

\begin{figure}[t]
  \centering
  \includegraphics[width=0.5\textwidth]{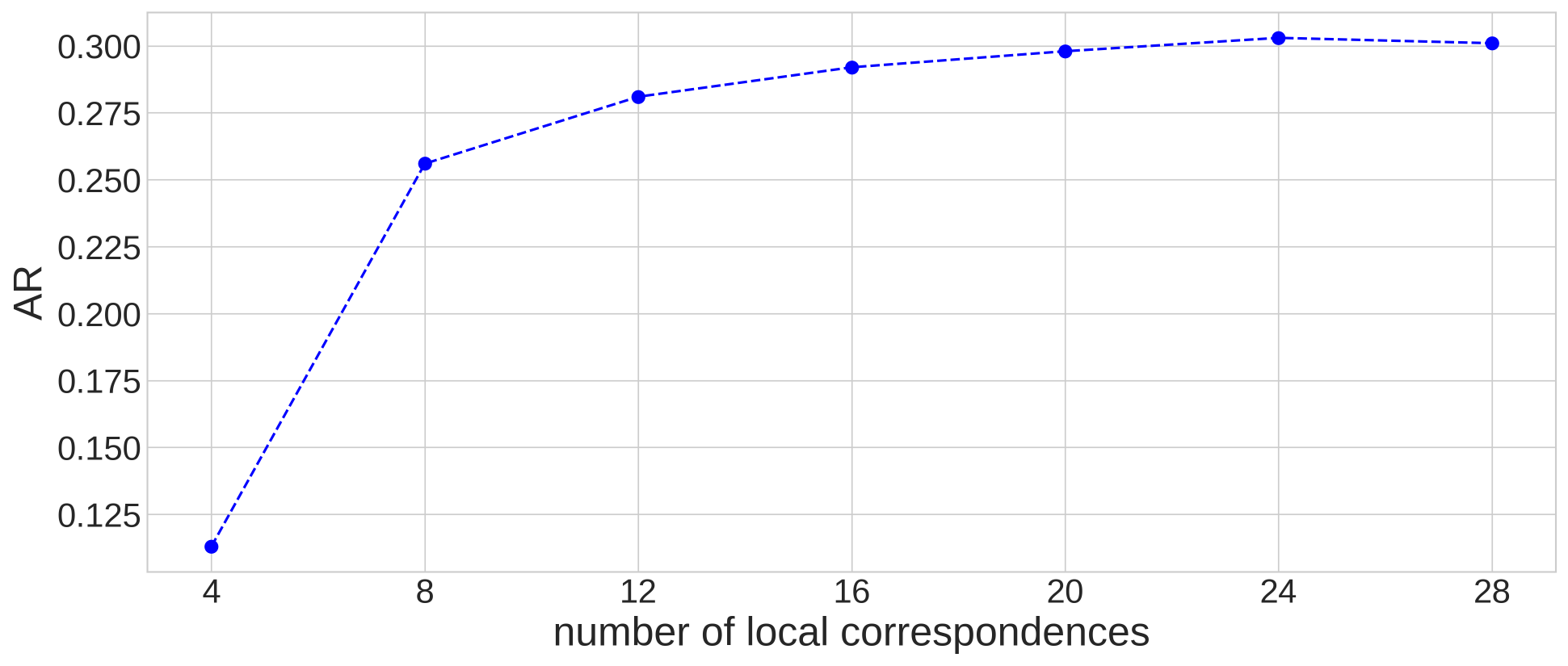}
  \caption{\textbf{Number of correspondences} Impact of the number of extracted local correspondences. Reported is the $AR$ score on LMO.}
  \label{fig:number_corrs}
\end{figure}

\subsubsection{Number of Correspondences}
Figure~\ref{fig:number_corrs} ablates the number of local correspondences used for deriving object coordinates.
Results are provided on LMO using the $AR$ score as a validation metric.
Considering the number of extracted local correspondences we observe a very similar behavior to the number of templates. 
The \(AR\) score rapidly increases with the number of local correspondences and flattens out around $20$ correspondences.
A higher number of local correspondences is not always feasible, due to the constraints enforced by Equation~\ref{eq:local_corr_calc}.
Additionally, using more correspondences increases the likelihood of wrong matches. 
This is partially compensated by the RANSAC~\cite{fischler1981random} iterations.

%\begin{figure}[h]
%    \centering
%    \begin{minipage}[b]{0.22\textwidth}
%        \includegraphics[width=\textwidth]{images/4_tless_in_corr.png}
%        \caption{Segmented object proposal with color-coded local correspondences.}
%    \end{minipage}
%    \hfill % This adds some space between the two minipages
%    \begin{minipage}[b]{0.22\textwidth}
%        \includegraphics[width=\textwidth]{images/4_tless_temp_corr.png}
%        \caption{Matched object template with corresponding color-coded local correspondences.}
%    \end{minipage}
%    \label{fig:correspondences_tless}
%\end{figure}

%\FloatBarrier

\section{Conclusions} \label{sec:conclusion}
We propose a zero-shot $6D$ object pose estimation method, which does not rely on task-specific fine-tuning and enables estimating poses of unseen objects.
The presented evaluations show that foundational Computer Vision models, precisely self-supervised ViTs are well-suited for extracting general image descriptors, and as such enable pose retrieval.
To be precise, we present results on LMO, YCBV, and TLESS, where we show that we can improve over results obtained by task-specific fine-tuned CNNs.
The current work focuses on generating initial pose hypotheses, without applying a refinement stage.
Future work will thus investigate how to refine pose hypotheses in a zero-shot fashion.

%%%%%%%%%%%%%%%%%%%%%%%%%%%%%%%%%%%%%%%%%%%%%%%%%%%%%%%%%%%%%%%%%%%%%%%%%%%%%%%%
% \addtolength{\textheight}{-8cm}   % This command serves to balance the column lengths

\bibliographystyle{IEEEtran}
\bibliography{root}

\end{document}